\documentclass[acmsmall,screen,authorversion,nonacm]{acmart}
\AtBeginDocument{%
  }

\setcopyright{acmlicensed}
\copyrightyear{2018}
\acmYear{2018}
\acmDOI{XXXXXXX.XXXXXXX}
\acmConference[Conference acronym 'XX]{Make sure to enter the correct
  conference title from your rights confirmation email}{June 03--05,
  2018}{Woodstock, NY}
\acmISBN{978-1-4503-XXXX-X/2018/06}

\begin{document}

\title{Eye-Tracking-Driven Control in Daily Task Assistance for Assistive Robotic Arms}

\author{Anke Fischer-Janzen}
\email{anke.fischer-janzen@hs-offenburg.de}
\orcid{0000-0003-3992-3163}
\affiliation{%
  \institution{Offenburg University of Applied Sciences}
  \city{Offenburg}
  \country{Germany}
}

\author{Thomas M. Wendt}
\affiliation{%
  \institution{Offenburg University of Applied Sciences}
  \city{Offenburg}
  \country{Germany}
}
\email{thomas.wendt@hs-offenburg.de}

\author{Kristof Van Laerhoven}
\affiliation{%
  \institution{University of Siegen}
  \streetaddress{Hölderlinstraße 3}
  \city{Siegen}
  \country{Germany}}
\email{kvl@eti.uni-siegen.de}

\renewcommand{\shortauthors}{Fischer-Janzen et al.}

\begin{abstract}
  Shared control improves Human-Robot Interaction by reducing the user's workload and increasing the robot's autonomy. It allows robots to perform tasks under the user's supervision. Current eye-tracking-driven approaches face several challenges. These include accuracy issues in 3D gaze estimation and difficulty interpreting gaze when differentiating between multiple tasks. We present an eye-tracking-driven control framework, aimed at enabling individuals with severe physical disabilities to perform daily tasks independently. Our system uses task pictograms as fiducial markers combined with a feature matching approach that transmits data of the selected object to accomplish necessary task related measurements with an eye-in-hand configuration. This eye-tracking control does not require knowledge of the user's position in relation to the object. The framework correctly interpreted object and task selection in up to 97.9\% of measurements. Issues were found in the evaluation, that were improved and shared as lessons learned. The open-source framework can be adapted to new tasks and objects due to the integration of state-of-the-art object detection models. 
\end{abstract}


\keywords{Robotics, Eye-Tracking, Assistive device, Shared control, Object detection, Feature matching}

\maketitle


\section{Introduction}
Eye-tracking-driven controls for robots and robotic arms can enable people with severe physical disabilities to regain independence in their daily lives. The ability to complete daily tasks independently has been shown to improve quality of life \cite{Calvo2008}. These controls vary in terms of their level of autonomy and interface design \cite{Fischer-Janzen2024}. Various control approaches are being developed, including teleoperated systems and sophisticated methods that use gaze analysis to predict user intent and control the robot \cite{Admoni2016, Huang2016}. When gaze is used as an explicit control input, it is often used alongside dwell-time approaches, in which the user fixates on an object with their gaze to select and interact with it. Researchers have found that this behavior comes naturally to users, as people tend to first look at the object they want to interact with in everyday situations \cite{Belardinelli2024}. 

Approaches that use fixation as an input modality rely on the accurate measurement of gaze location. Although authors often claim that eye-tracking glasses are inaccurate, their accuracy and precision have improved since the early systems described in the work of Kim et al. \cite{Kim2001}. Interest has shifted to three-dimensional (3D) gaze estimation, which determines the point at the intersection of the gaze vector and the surface \cite{Tostado2016,Li2017,Cio2019, Yang2021}. Gaining additional depth information is beneficial because it allows for more reliable tracking of gaze on object artifacts, such as the handles of cups. Therefore, task intent can be inferred, particularly when evaluating gaze in scenes with multiple objects \cite{Koochaki2021, Admoni2016}. Moreover, additional depth information can directly instruct a robot to grasp an object \cite{Cao2025}.  In many approaches, object location is estimated in relation to the user's head. Therefore, drifts in head movement measurements and slippage of the glasses can lead to significant estimation deviation.

As presented, these approaches focus on selecting and completing daily tasks. Various works demonstrate single-task implementations, such as those presented in \cite{Huang2016,Li2017,Cio2019, Yang2021}. More recent works focus on implementing multiple tasks that can be accomplished with eye tracking and multimodal systems \cite{Chi2025, Zhang2025}. Hagengruber et al. introduced the EDAN system, an sEMG-controlled assistive robot, and demonstrated the necessity of performing multiple daily tasks continuously and robustly to benefit real-world users \cite{Hagengruber2025}. This was confirmed in findings of Farhadi et al., highlighting the importance of seamless and intuitive interaction \cite{Farhadi2025}. However, most works lack this continuous task selection.

These approaches are promising in that they represent new control designs with a high level of robot autonomy. However, robust transmission of information requires high calibration accuracy in 3D gaze estimation in order to distinguish between tasks. Incorrect predictions and misinterpretations of user intent can lead to challenging situations in human-robot interaction and reduce trust in controls when the robot performs the wrong task or selects the wrong object \cite{Fischerjanzen2025}. Furthermore, selecting from a multitude of daily tasks would enhance real-world usability.

To address these challenges, we propose an eye-tracking–driven control system that reduces reliance on high gaze-tracking accuracy by shifting object localization to a robot-mounted (eye-in-hand) camera. Tasks are selected via task pictograms representing fiducial markers. Fiducial markers have been shown to enable robust object detection in unconstrained environments \cite{Aronson2018}. We transmit object information by integrating a YOLOv12n object detection model and a feature detection and matching approach, FLANN-AKAZE. Additionally, we integrated task pictograms from  \cite{Fischer-Janzen2025B} to enable the system to distinguish between multiple tasks. We also implemented a fallback method to ensure the correct object selection. The open-source framework presented here uses only eye-tracking glasses and the robot-mounted camera for sensory input. Thus, it presents a suitable setup for controlling a wheelchair-mounted robotic arm.

Our main contributions are:
\begin{enumerate}
    \item A comparison of feature-matching algorithms tailored to assistive robotic arm use cases used for data transmission.
    \item An open-source eye-tracking–driven robot control framework based on object detection solving head movement issues by reducing robot control to an eye-in-hand configuration.
    \item An evaluation of the framework that reveals key challenges and future research directions.
\end{enumerate} 

This work is organized as follows. In Section \ref{sec:Related_work} works are presented covering the topics of eye-tracking-driven robot control and task accomplishment. Section \ref{sec:methodsII} describes the anticipated human-robot interaction and provides implementation details. Section \ref{sec:MethodsExploration} evaluates an appropriate feature detection and feature matching approach. Section \ref{sec:ResultsFrameworkDescription} presents the framework, and Section \ref{sec:ResultsEvaluation} its evaluation concerning the task selection success rate. In Section \ref{sec:discussion} the results are discussed and concluded in Section \ref{sec:conclusion}.  

\section{Related work}
\label{sec:Related_work}
Recently, the development of eye-tracking-driven controls has been strongly influenced by three factors: 1) user intent prediction, 2) 3D gaze estimation, and 3) AI integration. As stated in the introduction, insights from these research fields improve the functionality and usability of using gaze as an input modality. However, this section will elaborate on the challenges that have been encountered. 

Shared control mechanisms aim to balance the interaction between users and autonomous robotic systems, ensuring accurate and efficient task execution. These mechanisms enable robots to perform tasks requiring fine motor skills, improve safety by constraining robot movements, and alleviate user workload \cite{Kim2012, Loke2025, Selvaggio2021, Cao2025}. Correct transmission of information between the eye-tracking device and the robotic system is crucial for the functionality of such controls. High error rates can cause the robot to interact with an object that the user did not select, resulting in increased frustration levels \cite{Farhadi2025, Bhattacharjee2020}. The necessary sensory information is defined by the robot control strategy and the level of autonomy. The level of autonomy describes the robot's abilities and determines the user's role in the interaction \cite{Beer2014}. These levels range from teleoperated robots that use gaze gestures or directional gaze to fully autonomous systems that, among others, use gaze to point to the desired object  \cite{Fischer-Janzen2024}. n this context, eye tracking is also used to increase the data density of brain-computer interfaces (BCIs), in which gaze is used to select objects and BCI input triggers the task \cite{Onose2012, McMullen2014}.

Accurately estimating a user's gaze in three-dimensional space is essential for precise object selection and interaction in complex environments. One approach is to detect an object's location by tracking the user's gaze in a 3D scene to select the object. Several authors have implemented 3D gaze estimation in robotic applications using various techniques. For example, Cio et al. used ray tracing to determine the intersection between the camera scene and the gaze direction \cite{Cio2019}. Li et al. calculated gaze location using gaze vectors and applied neural networks with pre-trained mapping relationships to reduce Cartesian errors \cite{Li2017}. Tostado et al. achieved 3D gaze tracking by developing a new calibration method that integrates the robot's trajectory to generate a continuous stream of gaze points in 3D space. They applied Gaussian processes for supervised learning \cite{Tostado2016}. Yang et al. augmented eye-tracking glasses with a camera that provides depth information. This information is used to calculate the intersection of the 2D gaze data and depth information. AprilTags were applied to enable head movement coordinate transformation, allowing users to move their heads \cite{Yang2021}. hese approaches rely on detecting the head's position in relation to the robot to interact with objects. In this context, fiducial markers are used to define workspaces or mark objects, thereby increasing accuracy \cite{Huang2016, Admoni2016,Aronson2020, Yang2021,Zhang2025}. In some cases, the eye-tracking glasses are equipped with additional sensors, such as IMUs or stereo cameras, which enhance the weight of the eye tracker \cite{Fan2020, Cao2025} and can lead to discomfort when worn for extended periods. An eye-in-hand configuration can address this issue by providing a known kinematic chain to the camera. 

Understanding and predicting user intent from gaze behavior is pivotal for making eye-tracking controls more intuitive and responsive. The natural tendency to gaze at an object when wanting to interact with it can be leveraged for this purpose \cite{Admoni2017, Belardinelli2024}. Belardinelli's work summarizes prediction models, including approaches using support vector machines, dynamic Bayesian networks, hidden Markov models, and others \cite{Belardinelli2024}. They applied this technique to mixing a drink, using fixation duration and gaze pattern sequence to predict the next ingredient \cite{Huang2016}. Yang et al. presented an approach that uses natural gaze and applies a target-attracted gaze movement model with a Kalman filter to extract features in object- and task-related gaze patterns \cite{Yang2023}. Gonzaléz-Díaz et al. investigated grasp prediction in multi-task learning involving tasks with multiple object interactions \cite{Gonzalez-Diaz2024}. Predicting user intent benefits robotics by providing necessary task and goal information \cite{Admoni2016}. However, the accuracy of these predictions is limited, leading to challenges when scaling this approach to various tasks or when the user goal changes unpredictably \cite{Belardinelli2024, Admoni2016}. Gaze behavior, such as peripheral vision, can lead to unintended fixations \cite{Aronson2018}. Differences in user behavior complicate the training of user intent prediction models furthermore \cite{Farhadi2025}.

Artificial intelligence techniques play a critical role in interpreting gaze data and enabling robots to effectively accomplish tasks. Gaze can be used in multiple ways to combine AI and robotics, as demonstrated by the work of Zhang et al. \cite{Zhang2020}. Previously, approaches were presented while elaborating on user intent estimation. Others use gaze to select objects, as demonstrated by Ivorra et al. in the AIDE European project, which employed electrooculography (EOG) and electroencephalography (EEG) to estimate gaze poses. YOLOv2 (You Only Look Once) was applied to select objects in the scene \cite{Ivorra2018}. Lastly, Large Language models are used to control eye-tracking robots by translating labels of selected objects into prompts for task execution \cite{Zhang2025}. 

Compared to the presented works, this approach differs in the following ways:
\begin{itemize}
    \item The framework's sensory inputs are received by the wearable eye-tracking glasses as the sole input modality and one additional low-cost camera mounted to the robot.
    \item The user can interact with objects in the real world without needing a GUI or display.
    \item No 3D-estimation of the user's position is needed, which allows for free head movement and reduces the influence of calibration errors due to the eye-in-hand configuration.
    \item The approach uses task pictograms as fiducial markers, which allow for robust selection between multiple tasks and are easy for users to understand.
\end{itemize}

\section{Interaction and implementation}
\label{sec:methodsII}
\begin{figure}[h] 
  \centering
  \includegraphics[width=\linewidth]{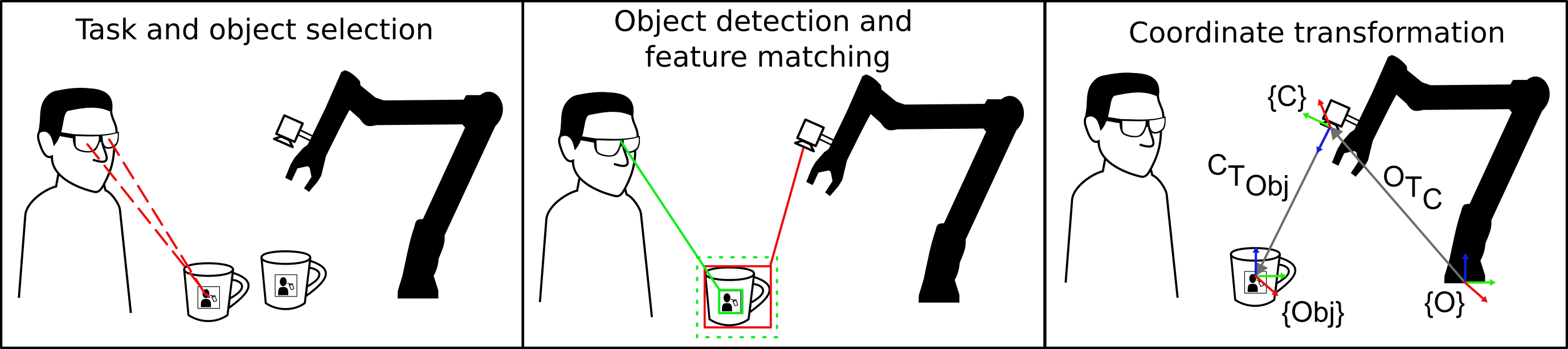}
  \caption{The computation process of the framework. The object is selected by gaze. The pictogram and the object are detected in both camera scenes and matched using a feature-matching approach. Coordinate transformation provides an estimated location of the object for further interaction.} 
  \Description{The computation process of the framework. The object is selected by gaze. The pictogram and the object are detected in both camera scenes and matched using a feature-matching approach. Coordinate transformation provides an estimated location of the object for further interaction.} 
  \label{fig:fig_function}
\end{figure}

This open-source, standalone, eye-tracking-driven robot control framework was designed to provide a foundation for integrating various robot-assisted daily tasks. To minimize accuracy errors caused by head movements, object data collection is transferred to the robot camera. This contrasts with current approaches, which use either a stationary camera or an eye-in-hand configuration. Figure \ref{fig:fig_function}  shows the general control process and computation steps that will be presented in Section \ref{sec:ResultsFrameworkDescription}. This process is presented in Section \ref{subsec:methods_hri}, which provides the rationale behind the control design choices. 

\subsection{Human-Robot Interaction process}
\label{subsec:methods_hri}
This control focuses on articulated robots that assist people with severe physical disabilities in accomplishing daily tasks, as shown in Figure \ref{fig:fig_function}. Gaze is the only input modality. Other input modalities that depend on extremity movement or speech are not integrated. Assistive robots are primarily implemented as an extension of electric or smart wheelchairs. They are referred to as Wheelchair-Mounted Robotic Arms (WMRA), Assistive Robotic Arms (ARA), or Physically Assistive Robots (PAR) \cite{Nanavati2024, Fischer-Janzen2024, Rahman2025}.
Due to close contact with users during activities such as assisting while eating and drinking, a predictable reaction of the robot has to be ensured. The goal of this work was to develop a reliable and robust interaction method that minimizes the need for additional peripheral devices (e.g. displays or stationary eye-tracking devices), acknowledging that people with severe physical disabilities already rely on multiple essential technologies such as speech computers or ventilators. More devices would complicate everyday usage and accessibility. Therefore, the presented framework requires only eye-tracking glasses and a RGB-D camera. 

The task selection is based on the approach presented in the work of Fischer-Janzen et al. (2025) that uses pictograms as fiducial markers \cite{Fischer-Janzen2025B}. The pictograms are used as visual cues and represent available daily tasks that are placed on related objects. For instance, the "Drink" pictogram is attached to a cup or glass. Users initiate autonomous task performance by fixating their gaze on the corresponding task pictogram. The object attached to the pictogram is defined as an object with which the robot can interact. One benefit of this method is that new tasks can be implemented quickly by expanding the pictogram dataset and retraining the detection model. A dwell-time approach was implemented, requiring the user to gaze at the pictogram for a certain amount of time. Due to the size of the pictograms, unintentional object selection is reduced compared to approaches that use whole objects to initiate interaction. 

With this control, the robot is expected to perform the task independently. This includes trajectory and grasp planning, as well as task execution. Thus, it integrates shared control approaches. The user's role is to select the object and task and then observe and correct the robot as it performs the task.

\subsection{Implementation details}
Current eye-tracking systems attempt to minimize selection errors by improving accuracy with 3D gaze estimation methods, as discussed in Section \ref{sec:Related_work}. The presented control system addresses this issue by transferring object localization and selection to a stationary or robotic arm-mounted camera (eye-in-hand configuration). This allows for the use of heavier, more accurate cameras than wearable eye-tracker scene cameras without increasing the weight of the eye-tracker itself. This eye-in-hand configuration is beneficial for WMRA applications.

The framework was implemented in ROS2 Humble running on Ubuntu 22.04. ROS2 Humble is a Robot Operating System (ROS) framework that provides multiple libraries for robot and sensor integration. It uses standardized topics and message types that facilitate communication between modules and improve matching the framework interface to other robots. Pupil Core glasses and Pupil Capture v3.5.1 (Pupil Labs \cite{Kassner2014}) were integrated to track eye movements. An Intel RealSense D455 publishes the robot’s field-of-view image data. ROS2 allows robots such as the Kinova Gen3 to be integrated using the manufacturer's repositories. Since this work focuses on general control that can be used to integrate various tasks and robots, Section \ref{subsec:ResultsSharedAutonomy} will discuss the implementation of robots and tasks.

\section{Method exploration}
\label{sec:MethodsExploration}
Transmitting data from eye-tracker to robot camera is done with feature matching. A comparison between algorithm was performed to find the best solution for the control. In this section we present the algorithm analysis. 

The OpenCV library offers established feature-detector-descriptors (FDD) and feature matching (FM) algorithms. However, the available options were limited by the version of OpenCV (version 4.5.4) used by other modules to match image variables between object detection models and ROS2 messages. The following feature detectors and descriptors are available: Accelerated-KAZE (AKAZE) \cite{Alcantarilla2013}, KAZE \cite{Alcantarilla2012}, Binary Robust Invariant Scalable Keypoints (BRISK) \cite{Leutenegger2011}, Oriented FAST and Rotated BRIEF (ORB) \cite{Rublee2011}, Scale-Invariant Feature Transform (SIFT) \cite{Lowe2004}, Speeded Up Robust Features (SURF) \cite{Bay2008}, Adaptive and Generic Accelerated Segment Test (AGAST) \cite{Mair2010}, and Maximally Stable Extremal Regions (MSER) \cite{Matas2004}. In the following, the feature matching algorithms Brute-Force Matcher (BF) and FLANN-based Matcher (FLANN) are evaluated. The algorithm implementation used to compare each algorithm was based on the tutorial provided by OpenCV \cite{OpenCV2025}. 

The evaluation was based on the idea that the object selected in the eye-tracker scene is the region with the greatest number of matched features in the robot scene. The bounding box generated in the robot scene that is closest to this feature cluster is most likely to represent the desired object. After the initial testing presented in Section \ref{subsec:method_in_the_wild}, the approach was refined to improve computation time and accuracy. This refinement involved comparing the bounding boxes of the eye-tracker camera scene with the bounding boxes of the robot camera scene. The number of features that are matched between two bounding boxes indicates the degree of object matching. 

\subsection{Comparison with literature}
A literature search was performed that included four well-established or recent publications covering the comparison of OpenCV based feature detection and matching algorithms \cite{Tareen2018, Noble2016, Pusztai2016, Golovnin2021}. Based on the insights gained from these publications, differences between versions and comparison methods were identified. Algorithms were excluded from further testing based on the authors' recommendations. Details are stated in Table\,\ref{tab:tab_OpenCV}. 

Some of the sources are ten years old, which limits their comparability with newer versions of OpenCV. This is especially true since different versions of OpenCV use different underlying programming languages. The available algorithms varied between versions. For instance, the SIFT patent expired in 2020, so it was excluded from older studies. Comparability was limited within some works due to the small number of detected features in ORB and GFTT, which were capped at 500 and 1,000, respectively. \cite{Golovnin2021, Pusztai2016}. Therefore, comparing performance, especially for feature detection and matching time, might be biased. The works of Golovnin and Rybnikov \cite{Golovnin2021} as well as Pusztai and Hajder \cite{Pusztai2016} found that FLANN-based FM performed worse than brute-force matching. Furthermore, both sources indicated that combining feature detectors and FMs can drastically affect performance, suggesting that feature detectors must be selected based on the FM.

The varying results led to the decision to repeat the tests using custom test cases designed for the robot application. This will allow to evaluate the performance of the FDDs and FM algorithms.

\begin{table}
  \caption{A comparison of publications on FDDs and FM algorithms based on OpenCV.}
  \label{tab:tab_OpenCV}
  \begin{tabular}{p{2.5cm}p{2.5cm}p{4cm}p{4cm}}
    \toprule
    Source & OpenCV Version & Dataset & Recommendation\\
    \midrule
    Tareen and Saleem (2018)\cite{Tareen2018} & OpenCV 3.3, Matlab 2017a & Images from University of OXFORD, MATLAB, VLFeat, OpenCV & SIFT, BRISK for highest accuracy and ORB, BRISK for highest efficency.\\
    Noble (2016) \cite{Noble2016} & OpenCV 3.1, C++& Benchmark image set "sculpture" & BRISK-BF for highest efficiency, SIFT-FLANN for most features matched.\\
    Pusztai and Hajder (2016) \cite{Pusztai2016} & OpenCV 3.0& Self-made Dataset & SURF, KAZE, AKAZE depending on images for highest accuracy.\\
    Golovnin and Rybnikov (2021) \cite{Golovnin2021} & OpenCV 4.5.1 Python & Lund University's motion pipeline data set & ORB-FLANN\\
  \bottomrule
\end{tabular}
\end{table}

\begin{figure}[h]
  \centering
  \includegraphics[width=0.67\linewidth]{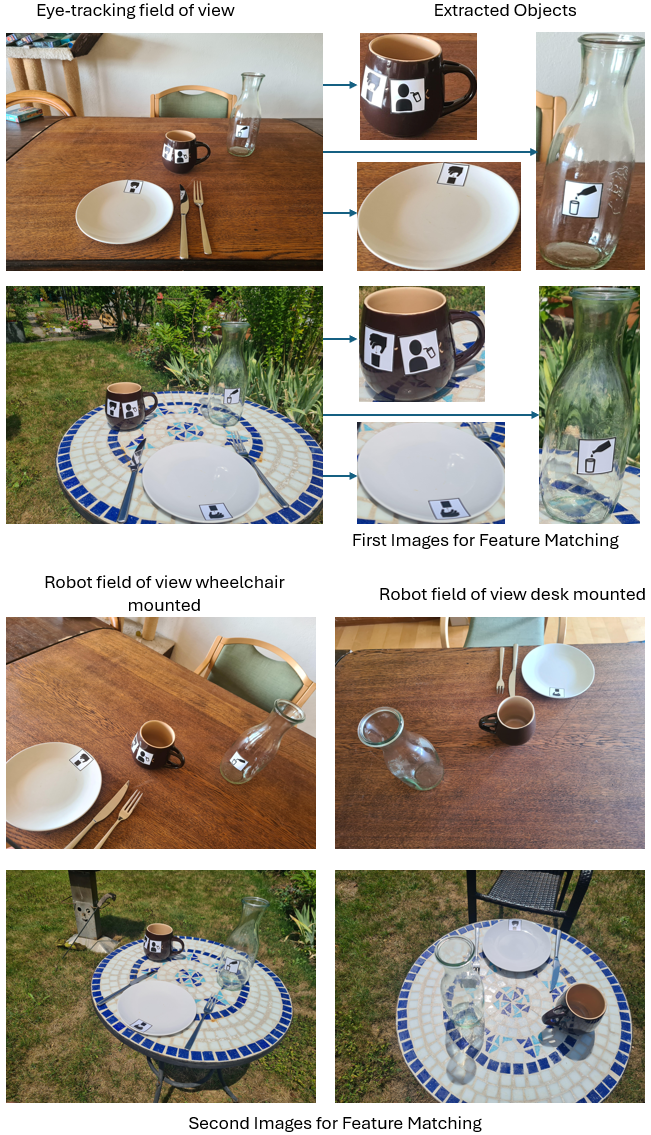}
  \caption{FM test cases. Upper scene: Inside scene of a set table. Lower scene: Outside scene of a set table in direct sunlight.} 
  \Description{Feature matching test cases. Upper scene: Inside scene of a set table. Lower scene: Outside scene of a set table in direct sunlight. The extracted objects are shown for each scene. Additionally, two images showing the fields of view of different robot applications are presented.}
  \label{fig:fig_testcases}
\end{figure}

\subsection{In-the-wild search approach}
\label{subsec:method_in_the_wild}
Two test cases were recorded and referred to as FM test cases. Both cases illustrate real-world scenarios, as demonstrated in Figure\,\ref{fig:fig_testcases}. The first test case depicts a set table in a home setting. The second case includes a table set up outside on a sunny day. These differences in lighting conditions and backgrounds were chosen to gain more detailed insight into the performance of the algorithms. From the field of view of the eye-tracker, bounding box cutouts are generated for the objects whose features will be used as descriptors in the FM algorithm. The image below simulates the field of view of a camera mounted on a robotic arm attached to an electric wheelchair. Most eye-in-hand configurations describe attaching the camera to the flange behind the end-effector. Therefore, cameras in this configuration are in an elevated position, closer to the scene than the user. In the indoor scene, the plate is partially outside the field of view, which matches real-world scenarios due to the natural misalignment between the user's view and the robot camera's view. In the second robot view, the camera is positioned on the robotic arm on the table across from the user. This configuration could be used for feeding tasks or to maximize separation of the shared workspace. Compared to standard feature-matching applications such as merging single photos into panoramas or video stabilization, this setup introduces challenges such as feature occlusion and misaligned edges. However, there are applications, such as feeding robots placed on the table, that require this setup.

The OpenCV 4.5.4 2D Feature Framework contains the following feature detection and description classes: ASIFT, AGAST, AKAZE, BRISK, FAST, GFTT, KAZE, MSER, ORB, and SIFT, and descriptor matchers: Brute Force Matcher and FLANN-Based Matcher. The excluded algorithms were ASIFT, AGAST, FAST, and GFTT, as they function as a keypoint detector with a missing descriptor and are implemented in other algorithms, such as ORB. KAZE was excluded because AKAZE demonstrated superior performance in the presented literature. MSER was excluded because it has no sophisticated matcher \cite{Pusztai2016}. The included algorithms were therefore AKAZE, BRISK, ORB, and SIFT. Both FMs were used in the experiment and paired with all FDDs, resulting in eight algorithm pairs.

Each permutation of FDD and FM was tested with the FM test cases. This resulted in ten feature-detection-description instances for each individual image (cup, plate, bottle, robot on wheelchair, and robot on a desk in each inside and outside scene). The best algorithm was selected based on computation time per feature matched, total computation time, and number of matched features after applying a distance ratio test suggested in the tutorial of OpenCV \cite{OpenCV2025} and introduced by Lowe \cite{Lowe2004}. The mean and standard deviation (SD) of the parameters were calculated for each algorithm pair and are shown in Table\,\ref{tab:tab_AlgorithmResults}. 

\begin{table}
  \caption{Algorithm comparison results. All values represent the mean value calculated from each match per algorithm.}
  \label{tab:tab_AlgorithmResults}
  \begin{tabular}{cp{3.0cm}p{3.5cm}p{3.5cm}}
    \toprule
    Algorithm & Total computation time (sec), mean ($\pm$ SD) & Computation time/feature matched (ms), mean ($\pm$ SD)  & Features matched, mean ($\pm$ SD) \\
    \midrule
    BF-AKAZE    & 1,728 ($\pm$1.239)    &1,371 ($\pm$0.886) &22,25 ($\pm$22.52)	\\
    BF-BRISK    & 3,795	($\pm$2.296)    &3,776 ($\pm$2.286) &10,83 ($\pm$9.81)\\
    BF-ORB      & 0,407	($\pm$0.455)    &0,326 ($\pm$0.108) &12,5 ($\pm$10.33)\\
    BF-SIFT     & 12,40	($\pm$3.109)    &12,243 ($\pm$2.997) &23,41 ($\pm$12.52)\\
    FLANN-AKAZE & 0,551 ($\pm$0.477)    &0,194 ($\pm$0.124) &22,83 ($\pm$23.26)\\
    FLANN-BRISK & 0,537	($\pm$0.349)    &0,517 ($\pm$0.339) &11,83 ($\pm$9.91)\\
    FLANN-ORB   & 0,377	($\pm$0.356)    &0,003 ($\pm$0.009) &17 ($\pm$16.75)\\
    FLANN-SIFT  & 1,229 ($\pm$0.548)    &1,068 ($\pm$0.437) &35,08 ($\pm$15.86)\\
     \bottomrule
\end{tabular}
\end{table}

\begin{figure}[h]
  \centering
  \includegraphics[width=\linewidth]{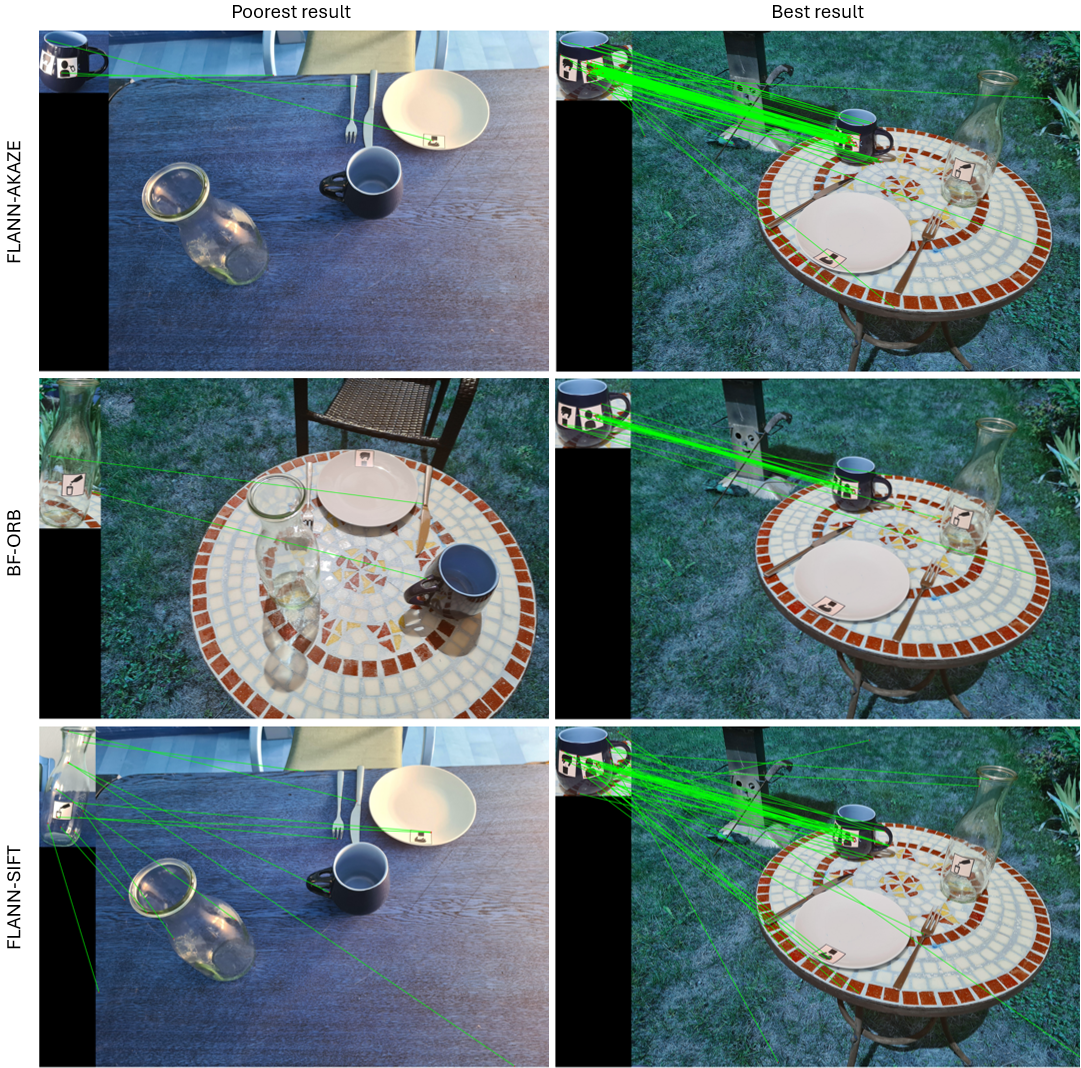}
  \caption{Feature matching results for selected algorithms. In the best cases, shown on the right 32 (BF-ORB), 87 (FLANN-AKAZE), and 53 (FLANN-SIFT) features were matched and not removed by the ratio test. In the worst cases, shown on the left 2 (BF-ORB), 3 (FLANN-AKAZE), and 12 (FLANN-SIFT) features were matched. Color changes in contrast to Figure \ref{fig:fig_testcases} result from using OpenCV library and have no impact on performance.} 
  \Description{Feature matching results for selected algorithms. In the best cases, shown on the right 32 (BF-ORB), 87 (FLANN-AKAZE), and 53 (FLANN-SIFT) features were matched and not removed by the ratio test. In the worst cases, shown on the left 2 (BF-ORB), 3 (FLANN-AKAZE), and 12 (FLANN-SIFT) features were matched.} 
  \label{fig:fig_FM_Auswertung1}
\end{figure}

The results from this test and literature were compared. It was confirmed that the fast computation time of ORB is due to the reduced number of 500 detected features, as stated in \cite{Pusztai2016, Golovnin2021}. Regarding the computation time for each feature detected, BRISK (0.019\,ms/feature), SIFT (0,16\,ms/feature) and AKAZE (0,35\,ms/feature) were faster than ORB (0.37\,ms/feature) as standalone FDDs. As highlighted in Table\,\ref{tab:tab_AlgorithmResults}, both BF-ORB and FLANN-ORB were the fastest algorithms in terms of total computation time, showing the stated computation time difference between FDDs and FDD-FM-pairs, and confirming the results of Golovnin and Rybnikov. This is due to the low computation time per matched feature in BF-ORB, which outperforms the other algorithms. SIFT with both FMs had the highest number of matches, followed by FLANN-AKAZE, BF-AKAZE, and FLANN-ORB. As indicated in Table \ref{tab:tab_OpenCV}, results of Noble et al. were confirmed for SIFT-FLANN, yet BRISK-BF is now one of the slowest algorithms. 

Based on these initial results the performance of FLANN-SIFT due to the high number of features matched; BF-ORB, due to its speed; and FLANN-AKAZE, due to good performances in both parameters were further investigated. The matched features were visualized for each algorithm as illustrated in Figure\,\ref{fig:fig_FM_Auswertung1}. It is evident in FLANN-AKAZE and FLANN-ORB that both algorithms detect the edges of the pictograms well. However, FLANN-SIFT detects and matches blobs, resulting in more diffuse matching. In the scenario with the robot mounted on the desk, the pictograms are occluded, resulting in worse prediction for all algorithms. FLANN-SIFT had the highest number of outliers in the best and worst case.

In conclusion, the in-the-wild search approach uses FLANN-AKAZE due to the trade-off between the large number of matched features and total computation time, as well as the relatively small number of outliers.

\subsection{Comparative approach} 
\label{subsec:Method_comparative_approach}
As presented, the accuracy of correctly matched features can vary drastically with robot position. The comparative approach was chosen to exclude areas that are not detected as objects, such as the tables in Figure \ref{fig:fig_FM_Auswertung1}. Matching the bounding box cutouts from both camera views was expected to yield the highest feature-match score for the pair depicting the same object. Due to the reduced image size searched for features, this method greatly reduces the computation time needed for algorithms such as SIFT.

A pairwise comparison of this approach was performed with BF-ORB, FLANN-AKAZE, and FLANN-SIFT due to their previous performance. The 12 instances observed included matching features in the user's view cutouts of a cup, bottle, and plate in scenes combined with the objects cutouts of the scenes indoor-desk, indoor-wheelchair-mounted, outdoor-desk, and outdoor-wheelchair-mounted. The ratio was set to different levels 0.6, 0.65, and 0.75 to estimate the occurrence of false positives and number of matched features. In the case of 0.6 and 0.65 ratio ORB did not find any matches in 3 of 12 instances, making it unusable for this approach. Most features were removed from the ratio test due to a high number of false matches for SIFT, with all ratios. Based on this outcome, FLANN-AKAZE was implemented in this approach. FLANN-AKAZE predicted the object correctly in 11 out of 12 instances, showing the highest number of matches. Changes in the ratio did not influence performance. In the outside scenario, it correctly matched objects more often than the other algorithms. Since it did not exhibit issues with zero values in feature matching, as ORB did, it was preferred. It was concluded that the algorithm is more robust in varying scenes and requires less fine-tuning.

\section{Framework description}  
\label{sec:ResultsFrameworkDescription}
In this section, we present the overall system framework, outlining the data flow from user interaction to robotic execution, that was introduced in Figure \ref{fig:fig_function}. Accordingly, in Section \ref{subsec:ResultsTaskSelector}, the communication with the eye-tracking device and pictogram detection used to realize a real-world gaze cursor are presented needed to initially select the object. In Section\,\ref{subsec:ResultsFeatureMatching} the transfer of data from eye-tracker to robot camera is discussed. The fallback mechanism to minimize the framework's arbitrary behavior and error sources incorporating both the in-the-wild search approach and the comparative approach is presented. This mechanism acknowledges the occurrence of misclassified bounding boxes and provides a reliable way to address such situations. Once the data has been successfully transmitted, the framework conceptually transitions to object interaction. Although no physical robot execution is included in this work, Section \ref{subsec:ResultsSharedAutonomy} describes the essential principles for estimating the object’s pose relative to a robot as a prerequisite for interaction.

\subsection{Eye-Tracking communication and task selection} 
\label{subsec:ResultsTaskSelector}
The open-source framework is available in this link\footnote{Link to repository: \url{https://github.com/AnkeLinus/EyeRAC}}. 
The EyeTrackerCommunicator, TaskDetection, and GazeCursor modules are described in this section. Communication with the eye-tracker was established using the Pupil Core Capture app. This app was further used for calibration, which takes less than a minute and is the only hardware that needs calibration before using the framework. The API backbone provided by Pupil Labs was integrated to ensure communication. In the EyeTrackerCommunicator, the "frame.world" and "fixation" topics were used to retrieve the necessary data published by two ROS2 Humble publishers created by the authors. The world video is published as an image message, and the fixation data is given as normalized x and y coordinates, with the lower left corner of the world video at 0/0 coordinates.

The GazeCursor logic employs a dwell-time approach, in which the user must maintain visual fixation on the object for a defined duration to confirm the selection. The API backbone automatically publishes an occurring gaze fixation, with a dwell time depending on the dispersion and duration of the detected fixation within the field of view \cite{PupilCore2025}. With pupil detection confidence greater than 0.95 at all times, the dwell time, and therefore the update rate of the publisher, is less than one second. Pictograms are detected and published in the TaskDetection module. The integrated model's training and augmentation used for pictogram detection is described in \cite{Fischer-Janzen2025B}. This model is trained on eight pictograms for the tasks "Drink", "Fill Cup", "Eat", "Scratch", "Switch Light Switch", "Brush", "Pick Object", and "Place Object".

The algorithm tracks new fixations and pictogram bounding box locations in relation to the eye-tracker scene camera frame. Overlap is calculated based on the locations. As soon as an overlap between a bounding box location and a fixation is found, the corresponding pictogram is selected. Then, a message is sent to the FeatureMatching node of type "Features". This message includes the selected bounding box information, such as location, width, height, and class, as well as task information and information about the existence of a secondary object. 

Reviewing the available tasks, it was evident that some require a secondary object to be completed. For example, the "Fill Cup" task requires at least one beverage and one cup or glass. The framework was enhanced with a secondary object selection. In these cases, the task is selected as usual, and then the user can select an object with or without an attached pictogram. This feature could be useful in social situations, such as a shared meal, allowing the user to select and fill guests' glasses.

\subsection{Human-Robot information transfer} 
\label{subsec:ResultsFeatureMatching}
\begin{figure}[h] 
  \centering
  \includegraphics[width=0.9\linewidth]{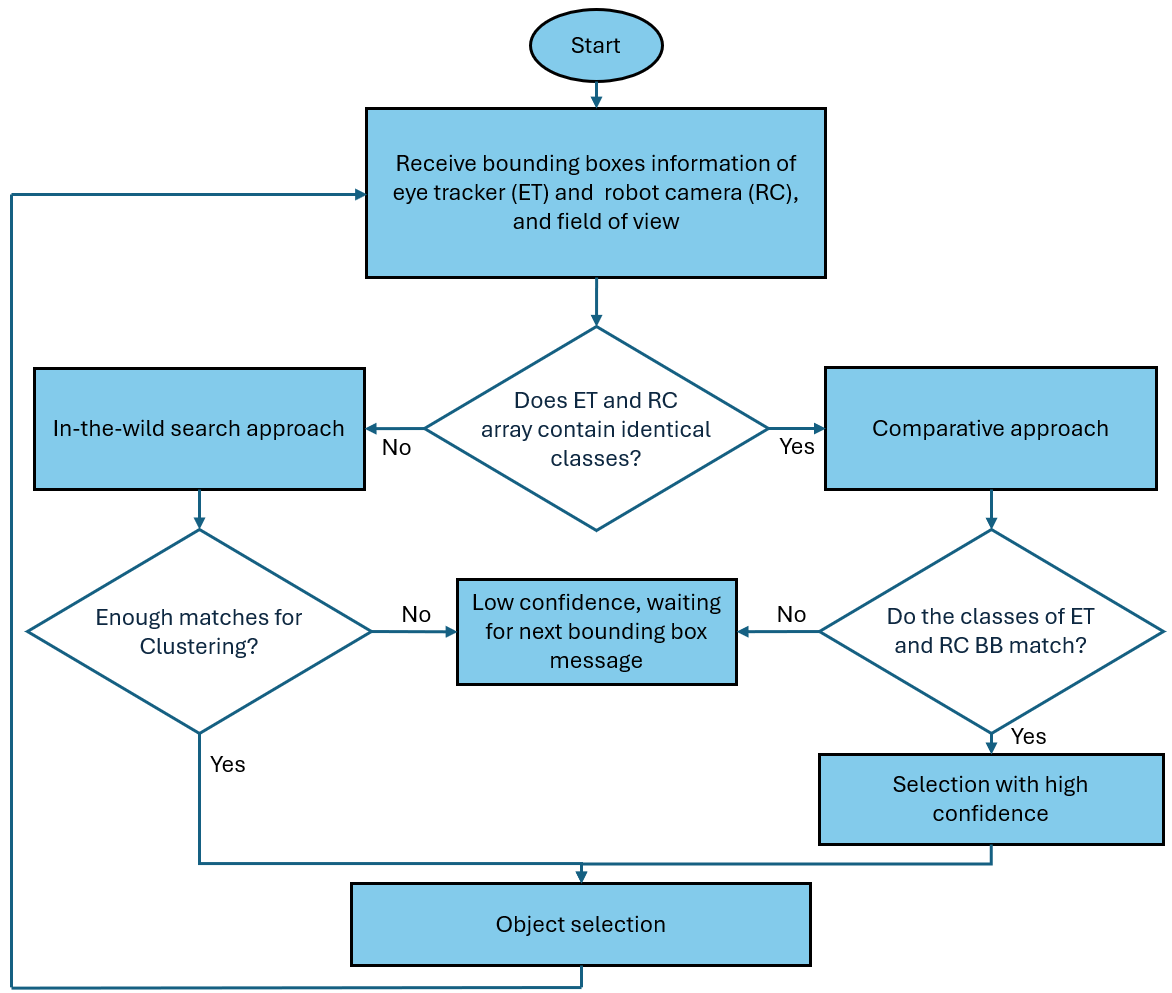}
  \caption{Program flow for robust selection information transmission between eye-tracker and robot field of view.}
  \Description{Program flow for robust selection information transmission between eye-tracker and robot field of view.} 
  \label{fig:fig_GC_to_FM}
\end{figure}

The user's selected object and the robot's interpretation has to be matched with high accuracy to ensure correct selection. To achieve this, the cutout of the pictogram is scaled up, so that the object is completely visible. Then both bounding box cutouts from the eye-tracker scene and from the robot camera scene are compared with FM. As seen the resuming falsly matched features still can lead to issues in object data transmission. Therefore, the aforementioned approaches were combined in a fallback mechanism to improve the selection process. The fallback mechanism is presented in Figure \ref{fig:fig_GC_to_FM} and checks if the bounding box class IDs match in both images. If this is true, there is a higher chance that object detection already indicates the selected object's existence. Hence, the comparative approach is chosen. Past experiments have indicated that, due to low representation in the training set, some classes may lack sufficient detection. For example, an object of the "cup" class may be detected as the "suitcase" class. In this case, the fallback mechanism clusters the features found in the complete robot camera image and calculates the distance to the closest bounding box. This behavior was realized with the in-the-wild search approach. The task and object IDs were matched using a custom library that converted the task class IDs into the object class IDs available in the MS COCO dataset and the assistive dataset \cite{Ponomarjova2025}.

Implementation of comparison approach: The bounding box cutout, representing the object attached to the selected pictogram is used to match detected features in all found bounding box cutouts. The cutout with the highest number of matching features will indicate the selected object for the robot. It is tested again to ensure that the object class of the final selected bounding box is identical to the necessary object for completing the selected task. As presented in the feature matching algorithm selection, the matching rate can be extremely low when the robot is mounted on the opposite side of the user (desk test case). Comparing the bounding box class can improve detection robustness and prevent incorrect predictions.

Implementation of in-the-wild search approach: If the class is not found in the robot view cutout set, then the robot field of view is used for feature matching. Clustering is then applied to identify the area with the highest density of matched features. The number of clusters is determined by the number of objects found in the scene, plus one additional cluster to account for outliers and missed bounding boxes. In this approach, k-means clustering was used to cluster matched features locations in the image of the robot field of view. The functionality of both approaches is evaluated in Section \ref{sec:ResultsEvaluation}.

\subsection{Shared control interface}
\label{subsec:ResultsSharedAutonomy}
ue to the reduction to an eye-in-hand configuration, coordinate transformation can be used to estimate the location of the object. Therefore, the accuracy is only limited by the errors of the robot, camera, and setup assembly. The coordinate transformation is illustrated in Figure \ref{fig:fig_function} and can be described as stated in Equation \ref{eq:1}, with the world coordinate frame $O$, the camera coordinate frame $C$ and the object coordinate frame $Obj$. $p$ describes the position of the corresponding coordinate frame in relation to the reference coordinate system.

\begin{equation}
\label{eq:1}
  {}^{O}p_{Obj}={}^{O}T_{C} \cdot {}^{C}p_{Obj}
\end{equation}

With ${}^{O}T_{C}$ describing the transformation from the world to the camera coordinate frame.

\begin{equation}
  {}^{O}T_{C} =
\begin{bmatrix}
{}^{O}R_{C} & {}^{O}p_{C}\\
\begin{matrix}
    0&0&0
\end{matrix} & 1
\end{bmatrix}	
\end{equation}

One suitable approach to obtaining model information from an object is to transform a 2D image cutout based on the bounding box into a 3D point cloud captured by the camera. However, since the Intel RealSense camera provides separate depth and color streams with different resolutions and principal points, depth pixels cannot be directly aligned with color pixels by index. Instead, depth values are reprojected into 3D using the depth intrinsics (Equation \ref{eq:3}), then transformed into the color camera reference frame through known extrinsics (Equation \ref{eq:4}), and finally reprojected into the color image using its own intrinsics (Equation \ref{eq:5}). A 3D point from the depth map is associated with the object when its corresponding color-image projection satisfies the bounding box inclusion condition defined in Equation \ref{eq:6}.

\begin{equation}
\label{eq:3}
\begin{bmatrix}
X_d \\ Y_d \\ Z_d
\end{bmatrix}
=
Z_d(u_d, v_d)\;
K_d^{-1}
\begin{bmatrix}
u_d \\ v_d \\ 1
\end{bmatrix}
\end{equation}

\begin{equation}
\label{eq:4}
\mathbf{P}_c
=
R_{dc}\, \mathbf{P}_d
+ \mathbf{t}_{dc}
\end{equation}

\begin{equation}
\label{eq:5}
\begin{bmatrix}
u_c \\ v_c \\ 1
\end{bmatrix}
=
K_c\;
\frac{1}{Z_c}
\begin{bmatrix}
X_c \\ Y_c \\ Z_c
\end{bmatrix}
\end{equation}

\begin{equation}
\label{eq:6}
u_{\min} \le u_c \le u_{\max},
\quad
v_{\min} \le v_c \le v_{\max}
\end{equation}

Libraries such as GPD (Grasp Pose Detection) \cite{Pas2017} can generate grasps based on an object's point cloud. Reducing the number of point cloud voxels representing the background can enhance the quality of the generated grasp. Preliminary tests showed that object segmentation methods produced adequate results. The framework's structure allows for the implementation of other tasks as well. The approaches are presented in Section \ref{subsec:DiscussionImplementationTasks}. 

\begin{figure}[h] 
  \centering
  \includegraphics[width=\linewidth]{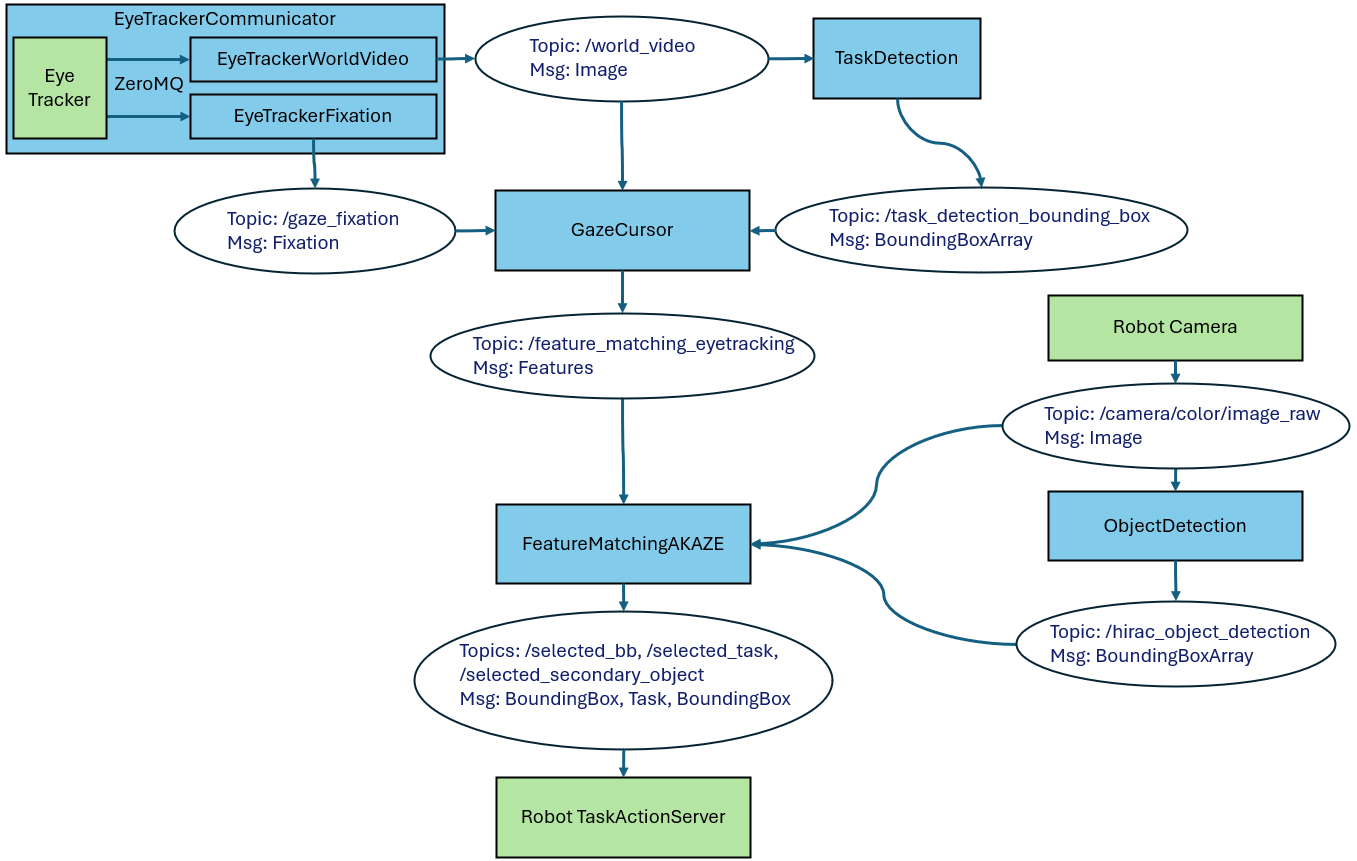}
  \caption{The framework of the eye-tracking controller. The depiction shows communication between nodes based on the ROS2 RQT graph. The rectangles represent nodes and the circles represent topics with message types. The green rectangles indicate interfaces to the periphery.} 
  \Description{The framework of the eye-tracking controller. The depiction shows communication between nodes based on the ROS2 RQT graph. The rectangles represent nodes and the circles represent topics with message types. The green rectangles indicate interfaces to the periphery. The nodes are the Eye-tracker Communicator, which sends data to the Task Detection, and Gaze Cursor nodes. The gaze cursor sends information to the feature matching node, which also receives information from the robot camera and object detection. After feature matching, the information is sent to the Robot task action server.} 
  \label{fig:fig_Framework}
\end{figure}

The resulting framework, including the introduced modules, is visualized in Figure \ref{fig:fig_Framework} to facilitate usability for other researchers.

\section{Evaluation}
\label{sec:ResultsEvaluation}
The evaluation focused on the reliability of object selection and data transfer. Specifically, we report a task selection success rate indicating how often the pictogram-associated object was correctly communicated to the robot side of the framework. A user-centered evaluation was not included, as comprehensive user satisfaction assessments inherently depend on additional components such as robot behavior, interaction dynamics, response timing and safety measures, which are not implemented within the scope of this work. Therefore, a meaningful usability study would require the fully integrated shared control system, which is reserved for future work.

\subsection{Test environment}
The test environment was setup under laboratory conditions to ensure comparability and reproducibility of the data and cases. Auditory cues were added, one when a pictogram was successfully selected and one when the according message was published, to inform the users that they can proceed. This greatly facilitated interaction in the experiment. One participant was recruited from the authors team to conduct the experiment. The participant was asked to wear the eye-tracking glasses and calibration was performed. The participant was instructed to gaze at the pictogram until the auditive cue was heard, then looking briefly at a random location away from the pictogram, followed by looking at the next pictogram in a defined sequence. After activating the robot camera via the ROS2 launch file and starting the framework, the measurements began. Three cases were tested, referred to as SU cases (SetUp cases) to distinguish them from the test cases in the feature matching evaluation. Before each test case, the eye-tracker was calibrated. These three cases examined how the framework reacted to the following aspects.

\begin{itemize}
    \item \textbf{SU Case 1:} Is the object and task information transmitted correctly to the robot framework? 
    \item \textbf{SU Case 2:} How are situations with falsely labeled or detected tasks and objects handled?
    \item \textbf{SU Case 3:} Can the framework detect a varying scene between user and robot field of view? 
\end{itemize}

\begin{figure}[h]
  \centering
  \includegraphics[width=0.7\linewidth]{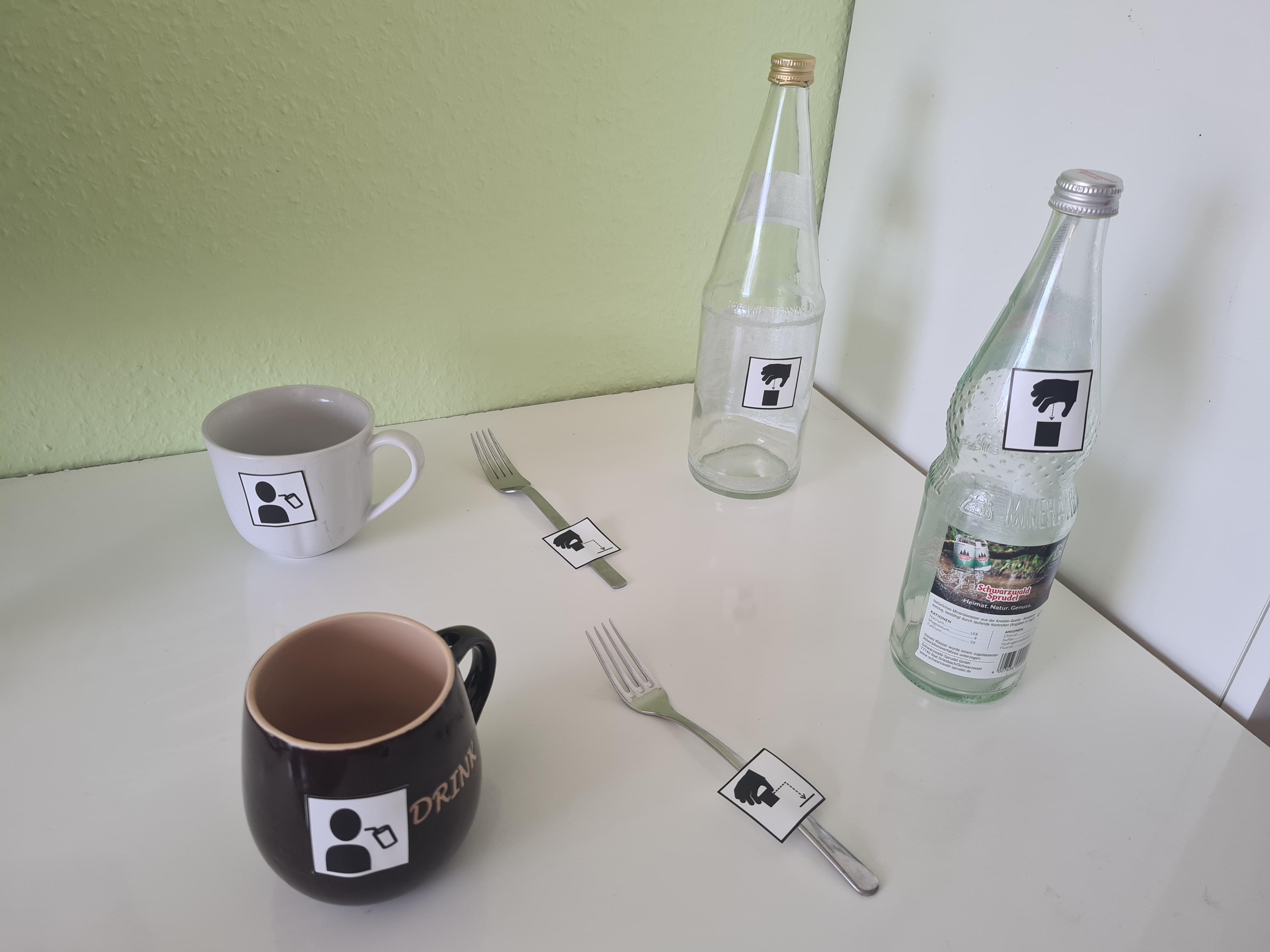}
  \caption{Test objects and attached pictograms. The upper row with white cup was used in all SU test cases. The lower row with brown cup was used for SU test case 3.1 and 3.2.} 
  \Description{Test objects and attached pictograms. The upper row with white cup was used in all SU test cases. The lower row with brown cup was used for SU test case 3.1 and 3.2.}
  \label{fig:fig_testcaseobjects}
\end{figure}

In all SU cases, the interactable objects were a cup, a fork, and a bottle as seen in Figure \ref{fig:fig_testcaseobjects}. The combination and tasks attached varied between cases as stated below.

SU Case 1: This case was used to represent perfect object detection and labeling. The cup, bottle, and fork in the upper row of Figure \ref{fig:fig_testcaseobjects} were placed in front of the slightly elevated robot camera. Correct object detection of all objects was ensured at any point in time by placing them at distinct points in the scene. The cup had attached the pictogram for task "Drink", the fork the task "Place", and the bottle the task "Pick". The user sat next to the robot camera, wearing the eye-tracking device, in approximately 80\,cm to 100\,cm distance to the objects. The comparison approach was tested with this setup.

SU Case 2: This condition was created to check how the system handles detected objects with wrong label. All task pictograms on the objects in Case 1 were replaced with the task "Place". Additionally, the class of the object in the task-to-object library was switched to a non-existent object class to ensure that in no case the object could be matched. This has lead always to the in-the-wild search approach. The participant was placed and behaved similarly to SU Case 1.

SU Case 3: False positives were evaluated. In this scenario, it was assumed that user and robot do not face the same scene, such as in the case of the user facing away from the robot to talk with someone or just observing the surroundings. In this case, all objects illustrated in Figure \ref{fig:fig_testcaseobjects} are used. The upper row of objects in the figure is placed in the robot's field of view, while the lower row is placed outside the robot's field of view and inside the user's. The SU case was conducted twice. First, all pictograms were attached, as shown in the figure. Second, the pictograms were removed from the robot scene to evaluate their influence on FM. The participant was seated in the same room but facing away from the robot's field of view to ensure they observed different scenes.

\subsection{Test parameters and data preparation}
The ratio was set to 0.75 for the comparison and in-the-wild search approach. The task selection success rate and mean task selection time were calculated. Since the selection process sometimes triggered multiple times due to their fast reaction time, duplicates were excluded from the measurements. In the following, the number of included measurements will be stated. In each SU case at least 100 measurements were recorded, each describing one task selection via gaze. Since the measurements and repetitions were counted by the participant, the number of measurements may vary from 100 due to human error. Additional data was recorded, such as time stamps and resulting features left after the ratio test, as well as cluster sizes. This data was used to look into error sources.

\subsection{Evaluation Results}

\textbf{SU Case 1:} The eye-tracking glasses had an accuracy of $0.545^\circ$ and a precision of $0.096^\circ$. In total, 156 measurements were recorded, of which 74 were included in the evaluation. A successful message was only sent to the robotic framework when the feature matching could determine the object with the maximum features matched in the comparison approach, and if the object classes matched for the selected object between user and robot scene. These limitations led to a task selection succes rate of 97.1\%, in which the data of the object and task selected by the user was correctly transmitted to the robot. The task selection time, from the selection of the object until the transmission of the message, was 260.3\,ms (SD 88.5\,ms).

\textbf{SU Case 2:} In this case, the objects' classes in the robot scene are not matching with the classes predicted from the eye-tracker. For example, task "Drink" assumes the existence of a cup, but "cup" classes are not present in the bounding box array of the robot scene. The eye-tracker had an accuracy error of $0.629^\circ$ and a precision error of $0.072^\circ$. 186 measurements were recorded. Excluding duplicates, 102 measurements were included in the evaluation. In 59 measurements (57.8\%), the message was sent, of which 35 messages predicted the bottle correctly, 23 predicted the fork correctly, and 1 message predicted the cup correctly. An assumption why information for cup was so sparsely sent might result in few matched features by FLANN-based matching. The ratio test removed all matches found in comparison to the other classes, even though the amount of matches found in all classes were similar, ranging between ten and twenty. Samples for all classes revealed that the locations of matched features for the cup scattered more in the image than for the other two objects. Since all pictograms are the same, the features matched with the pictograms are ambiguous. The features matched in the rest of the image are the deciding factor. The white cup on the white table might be confused with the transparency and reflections on the bottle or fork. Therefore, the bounding box classes mismatch and the message will not be sent. In the other 42.2\% of measurements no message was sent due to an insufficient number of matches. Yet, it is rather better to not send a message than a wrong message. This could lead to a unpredicted task completion, that might lead to user dissatisfaction and lost sense of agency.

\textbf{SU Case 3:} The first trial with attached pictograms in both camera scenes, 168 measurements were recorded with the eye-tracker accuracy of $0.629^\circ$ and precision of $0.072^\circ$. 95 measurements were included after removing duplicants in the evaluation, of which 45 messages were sent (47.4\%). This indicates that these measurements would result in a false positive and selection of the wrong object due to the separated field of view of user and robot. 

It was assumed that removing the pictograms from the robot set would reduce the false positive rate. The test was repeated, leading to 98 included measurements, of which 61 measurements were sent (57.4\%). The higher number of sent messages could result from varying lighting conditions influencing the feature matching, since the room was lit by a window and the measurements were not done at the same time of day. We will discuss the effects between SU test cases and the effects on interaction in Section \ref{subsec:discussionEvaluation}. 

\section{Discussion} 
\label{sec:discussion}
In the following, the found effects are discussed. These insights were used to explain possible optimization strategies to fine-tune the control. Lastly, the integration of task routines for robots are discussed. 

\subsection{Lessons learned}
\label{subsec:discussionEvaluation}
In summary, method exploration revealed that although SIFT has more matched features, the trade-off of fewer false positives and faster computation time yielded promising results. Consequently, the method was used in the comparison and in-the-wild approaches. Performance was tested in the framework evaluation. As stated in Table \ref{tab:tab_Evaluation_Sum}, the FLANN-AKAZE algorithm achieved adequate results. The performance of the eye-tracking-driven control was evaluated in different cases that investigated potential challenges of such a control. In Table \ref{tab:tab_Evaluation_Sum} the SU case results are summarized. 

Overall, the number of messages sent was low in all categories, ranging from 44.9\% to 57.8\%. Messages were not sent when objects were classified differently than expected in the eye-tracking task selection. For example, the "drink" task was selected, but a fork was identified as the most matching object in the robot camera scene. Differences between the approaches were found indicated by the task sent rate. SU cases 1 and 3.1 used the comparative approach, while SU case 2 used the in-the-wild search approach. SU case 3.2 also used the in-the-wild search approach in 31 out of 61 cases. As can be seen, the task sent rate is lower with the comparative approach, indicating stricter behavior that improves the task selection success rate. The low number of messages was considered a minor drawback, as the average task selection rate was 260.3\,ms per calculation. Since other studies have shown that acceptable dwell times can be twice this duration \cite{Holmqvist2017}, it is assumed that the waiting time for a repeated measurement is appropriate for users.  

With a mean task selection success rate of 95.67\%, incorrect predictions about the transmission of the object are rare. However, as tested in SU Case 3, this does not always result in good selection rates. Due to the experimental design, it was expected that the task sent rate would drop near zero in cases 3.1 and 3.2. There were fewer matched features compared to the other test cases. However, since the minimum threshold for matched features was set too low, the message was sent anyway. The threshold was implemented more strict in the code, resembling one lessen learned. 

By comparing the messages sent for each object, we deduced that the following aspects could influence the selection: Due to the test setup, it was assumed that each object was selected equally since participants had to follow a sequence when looking at the objects. Duplicates were removed before calculating the number of messages sent for each object class. In each case, one of the objects was correctly detected more often than the others. In SU cases 1, 2, and 3.2, the bottle was detected more often. As soon as the in-the-wild search approach is conducted, the bottle has a slight advantage of being selected since it has the biggest bounding box and covers the most space in the camera scene, as seen in SU cases 2 and 3.2. SU case 1 used the comparison approach. Due to the size of the bottle's bounding box, a larger cutout was created, allowing AKAZE to detect more features. This contrasts with SU case 3.1, in which the fork is the most correctly detected object and the pictogram leads to a high number of matched features. In SU case 3.1, where the robot and participant viewed scenes with varying objects, the fork was selected most of the time. One reason might be that the selected forks did not vary as much as the bottle and cup. Unlike in case 3.2, the pictograms were applied to both sets of objects, which may have increased the number of matched features.

\begin{table}
  \caption{Summary of the evaluation test cases, task selection success rates, and correct messages sent. Task sent rate refers to all measurements, while task selection success rate and messages sent for objects are calculated on the basis of the total number of messages sent.}
  \label{tab:tab_Evaluation_Sum}
  \begin{tabular}{cccccccc}
    \toprule
    Test case   &  Total    &  Messages  & Task send  & Task selection  & \multicolumn{3}{c}{Messages correctly send for} \\
                &  meas.          &  send          &  rate  &   success rate &  bottle & cup & fork \\
    \midrule
    SU Case 1      & 74            & 34            & 47.3\%    & 97.1\%     & 26 & 4  & 4 \\
    SU Case 2      & 102           & 59            & 57.8\%	   & 92.8\%    & 35 & 1  & 23 \\
    SU Case 3.1    & 95            & 45            & 47.4\%    & 97.9\%    & 9  & 10 & 26 \\
    SU Case 3.2    & 98            & 61            & 57.4\%    & 94.9\%    & 40 & 21 & 0 \\        
     \bottomrule
\end{tabular} 
\end{table}

As indicated, the potential advantages of pictograms in this feature matching approach compared to other selection methods were examined. In conclusion, it is evident that pictograms impact the feature matching algorithm. When comparing Case 3.1 (where the objects had identical pictograms) and Case 3.2 (where the pictograms were removed from one set of objects), the task selection rate was not significantly different; however, smaller objects, such as forks, were easier to detect. Facilitated feature detection with pictograms was also found when comparing the two robot positions in Figure \ref{fig:fig_FM_Auswertung1}.  Pictograms have edges and varying designs with high contrast that represent the features searched for by AKAZE. Different results may be obtained using other FDD algorithms.

The lessons learned from the evaluation can be summarized as the following three approaches that were designed and integrated into the control. Before using this framework researchers should apply these lessons to fit the framework to their use case.
\begin{enumerate}
    \item Calibrating the resizing of bounding box cutout sizes in robot scene
    \item Calibrating the threshold for minimum number of matched features
    \item Choosing appropriate ratio for ratio test
\end{enumerate}

The threshold and ratio were integrated as parameters into the FeatureMatching module and can be adjusted as needed. Online searches suggest that the ratio should be between 0.65 and 0.8. The limits of 0.6 to 0.75 were tested during method exploration. No significant difference was found. Since the Euclidean distance is determined by this ratio, 0.65 is the stricter limit, resulting in fewer matched features. However, due to the cameras' resolution and the blurriness caused by head movements, a less strict ratio, such as 0.75, is recommended. The threshold is, by default, set to five successfully matched features after the ratio test, which is chosen by the lower boundary in the successful send tasks.

\subsection{Implementation of robot task accomplishment}
\label{subsec:DiscussionImplementationTasks}
The implemented task detection model includes the following tasks: "drink," "fill cup," "eat," "scratch," "switch light switch," "brush," "pick object," and "place object." After the presented task selection, the according task has to be performed by he robot. Several research groups are developing solutions for each task. Due to the breadth of related work considering each task, the following discussion about suitable approaches is not exhaustive. We present solutions that demonstrate high-level robot autonomy in completing tasks. If more information is needed to accomplish a task, eye-tracking approaches are presented to continue the modality availability. Some works can perform multiple of the aforementioned tasks. For the sake of conciseness, they are only presented once.

\textbf{Drinking:} The Assistive Gym is a physics-based simulation framework for physical human-robot interaction and robotic assistance, using reinforcement learning to control the robot in various tasks. With this framework, tasks such as drinking can be performed, as demonstrated in Erickson et al. \cite{Erickson2020}. Due to the large number of integrated robots, this framework is a promising solution for integrating this control. Chi et al. presented a trajectory segmentation approach that can fill a glass of water, hold it near the user, and assist with eating tasks, representing three of the presented tasks \cite{Chi2025}. The method by which users can select among these tasks is not specified, underscoring the significance of our framework.

\textbf{Fill cup:} Huang and Mutlu's work presents a user intent prediction-based approach in which users can select individual ingredients via eye tracking to mix a drink. This setup has been used to demonstrate that anticipatory control is faster than reactive control \cite{Huang2016}. This approach could help expand possibilities for people with severe disabilities, for example new employment fields.

\textbf{Eating:} Bhattacharjee et al. examined user preferences in a robot-assisted feeding task. They introduced a system that could pick up specific food items via a web interface \cite{Bhattacharjee2020}. Through our implementation of the GazeCursor, interaction with displays and web interfaces is anticipated. Canal et al. presented a setup that can feed people and assist them with dressing themselves \cite{Canal2021}. 

\textbf{Switch Light Switch:} Zhang et al. presented a sophisticated, eye-tracking-based robotic approach capable of picking up and placing objects, toggling switches, watering plants, and pouring water. These tasks were selected based on prompts generated by an LLM model that uses object labels. This approach can be scaled to multiple tasks \cite{Zhang2025}. Our approach could still be beneficial by enhancing the data density of the given pictograms.

\textbf{Scratch and Brush:} Both Hughes et al. and Dennler et al. present an approach to combing hair \cite{Hughes2021, Dennler2021}. Although this approach is not based on eye-tracking control, the physics are discussed, including different hair types and lengths. No works covering itch-scratching with a robot were found. The literature covers stroking or touching humans as a form of calming social interaction. Scratching is most likely not covered due to the potential for user and participant harm, which is why this task is usually assigned to caregivers. As Asimov's first law of robotics states, "A robot may not injure a human being or, through inaction, allow a human being to come to harm." \cite{Asimov1950}.
 
\textbf{Pick and place:} These tasks are becoming established as proof of concept for eye-tracking controls. They lay the groundwork for the aforementioned approaches since most tasks involve reaching for and grasping objects. As presented in Section \ref{subsec:ResultsSharedAutonomy}, the theoretical description was given to obtain coordinates. These coordinates can be used to place an object by calculating the intersection of the surface and gaze vectors. GPD was also mentioned as a solution for generating appropriate grasps \cite{Pas2017}. We refer at this point to Marwan et al., who present a review including multiple eye-in-hand approaches for grasping \cite{Marwan2021}.

\subsection{Limitations and future work}
The system must be tested in everyday situations, including scenes with clutter and varying environmental parameters. The test cases were chosen to demonstrate various settings, but were limited to three specific tasks. We documented lessons learned to facilitate use by other researchers. Furthermore, the current integrated, assistive-dataset trained model can detect and track assistive objects with high confidence. However, it is limited to a few classes. The presented logic allows for the use of different models trained on broader datasets to integrate other objects for interaction. Thus, the model can be switched to YOLO models, which are trained on MS COCO and include approximately 80 different classes.

Regarding the use of the framework by people with severe disabilities, an adjustment measure for the robot's field of view should be integrated so that those unable to move their heads can independently adjust the view to align with their gaze. Multimodal inputs have shown promise in opening the framework to people who cannot interact with the provided input modality \cite{Onose2012, Atienza2005}. Furthermore, interaction and user satisfaction can be improved by refining feedback modalities. In this work, we used auditive cues to indicate successful selection. Other studies have integrated mixed reality, enabling visual cues in the user's field of view \cite{Park2022, Baptista2025}. dditionally, LLMs enable more sophisticated auditory feedback, such as information about the object or selected task. These refinements are planned for future work.

\section{Conclusion} 
\label{sec:conclusion}
In conclusion, this work introduces a novel eye-tracking-driven shared control framework for assistive robotic arms that effectively addresses challenges related to gaze accuracy in 3D-object detection and complexity related to robustly detecting correct task selection in a system with multiple tasks available. By leveraging task pictograms as fiducial markers and employing a robust feature matching-based communication between the user’s eye-tracker scene and a robot-mounted camera, the system enables reliable and intuitive interpretation of task selection without requiring head position tracking. The object's location is found by a kinematic chain through a eye-in-hand configuration. Experimental evaluations demonstrate a high task selection success rate of up to 97.\% with a mean response time of 260.3\,ms, underscoring the system’s robustness and efficiency. While the system reacted adequately in some test cases downfalls were discovered, leading to wrong system reaction. As reaction, counter measures, such as threshold for minimal detected features, were implemented in the code and presented as lessons learned. The open-source framework’s modular design and integration of state-of-the-art object detection models facilitate easy adaptation to new tasks and environments. While current tests are limited to laboratory settings and a finite set of objects is available, we presented ways to adapt the system to benefit new approaches. This approach lays a foundation for the integration of real-world applications, that can enhanced independence, usability, and safety for users with severe physical disabilities through autonomous robot assistance. 

\begin{acks}
We would like to thank the following contributors to the HIRAC project: Bastian Kayser, Katrin Ponomarjova, Tobias Krauss, Vanessa Stöckl, Madhurshalini Mahalingam, Ebi Eilber, and Cadida Software GmbH.
ChatGPT3.0 was used to improve the code for visualizing feature matching in Figure \ref{fig:fig_FM_Auswertung1}. Additionally, ChatGPT3.0 and DeepL Write were used to improve the grammar.
\end{acks}

\bibliographystyle{ACM-Reference-Format}
\bibliography{sample-base}

\end{document}